\documentclass[10pt,twocolumn,letterpaper]{article}

\usepackage{iccv}
\usepackage{times}
\usepackage{epsfig}
\usepackage{graphicx}
\usepackage{amsmath}
\usepackage{amssymb}

\usepackage{booktabs}
\usepackage[dvipsnames]{xcolor}
\usepackage{xcolor,colortbl}
\usepackage{color, colortbl}
\usepackage{xcolor}

\usepackage{sidecap}

\usepackage{float}
\definecolor{purple}{rgb}{0.65,0,0.65}
\definecolor{dark_green}{rgb}{0, 0.5, 0}
\definecolor{blueish}{rgb}{0.0, 0.3, .6}
\definecolor{lightgreen}{RGB}{238, 252, 241}
\definecolor{lightred}{RGB}{231, 187, 187}
\definecolor{darkred}{RGB}{198, 129, 129}

\definecolor{tabhighlight}{HTML}{e5e5e5}
\usepackage{graphicx, amsmath, amssymb, caption, subcaption, multirow, overpic}
\definecolor{tabhighlight}{HTML}{e5e5e5}
\definecolor{citecolor}{HTML}{0071bc}

\newcommand{\method}{ECO\xspace}

\usepackage[breaklinks=true,bookmarks=false]{hyperref}

\usepackage[capitalize]{cleveref}
\crefname{section}{Sec.}{Secs.}
\Crefname{section}{Section}{Sections}
\Crefname{table}{Table}{Tables}
\crefname{table}{Tab.}{Tabs.}

\iccvfinalcopy % *** Uncomment this line for the final submission

\def\iccvPaperID{****} % *** Enter the ICCV Paper ID here

\ificcvfinal\fi

\begin{document}

%%%%%%%%% TITLE
%\title{\LaTeX\ Author Guidelines for ICCV Proceedings}
\title{ECO: Ensembling Context Optimization for Vision-Language Models}
%\title{Learning to Prompt with Context Ensembles}
%\def\equalcontribution{*}\footnotetext{Equal contribution}

\author{Lorenzo Agnolucci\thanks{Equal contribution}\\
University of Florence\\
{\tt\small lorenzo.agnolucci@unifi.it}
\and
Alberto Baldrati\footnotemark[1]\\
University of Florence\\
{\tt\small alberto.baldrati@unifi.it}
\and
Francesco Todino\\
University of Florence\\
{\tt\small francesco.todino@stud.unifi.it}
\and
Federico Becattini\\
University of Siena\\
{\tt\small federico.becattini@unisi.it}
\and
Marco Bertini\\
University of Florence\\
{\tt\small marco.bertini@unifi.it}
\and
Alberto Del Bimbo\\
University of Florence\\
{\tt\small alberto.delbimbo@unifi.it}
}

\maketitle
\ificcvfinal\thispagestyle{empty}\fi

%%%%%%%%% ABSTRACT
\begin{abstract}
   Image recognition has recently witnessed a paradigm shift, where vision-language models are now used to perform few-shot classification based on textual prompts. Among these, the CLIP model has shown remarkable capabilities for zero-shot transfer by matching an image and a custom textual prompt in its latent space. This has paved the way for several works that focus on engineering or learning textual contexts for maximizing CLIP's classification capabilities. In this paper, we follow this trend by learning an ensemble of prompts for image classification. We show that learning diverse and possibly shorter contexts improves considerably and consistently the results rather than relying on a single trainable prompt. In particular, we report better few-shot capabilities with no additional cost at inference time. We demonstrate the capabilities of our approach on 11 different benchmarks.
\end{abstract}

%%%%%%%%% BODY TEXT
\section{Introduction}

Thanks to their large-scale pre-training, foundational vision-language models proved to be very effective at generalizing to downstream tasks. In particular, CLIP (Contrastive Language-Image Pre-training) \cite{radford2021learning} has achieved surprising performance in several different fields, such as image generation \cite{galatolo2021generating}, image retrieval \cite{baldrati2022effective,baldrati2023zero} and image quality assessment \cite{wang2023exploring}. Specifically, CLIP can be employed for zero-shot classification by predicting the output class based on the similarity between the image features and the textual features of words belonging to a given vocabulary.

\begin{figure}
    \centering
    \includegraphics[width=\linewidth]{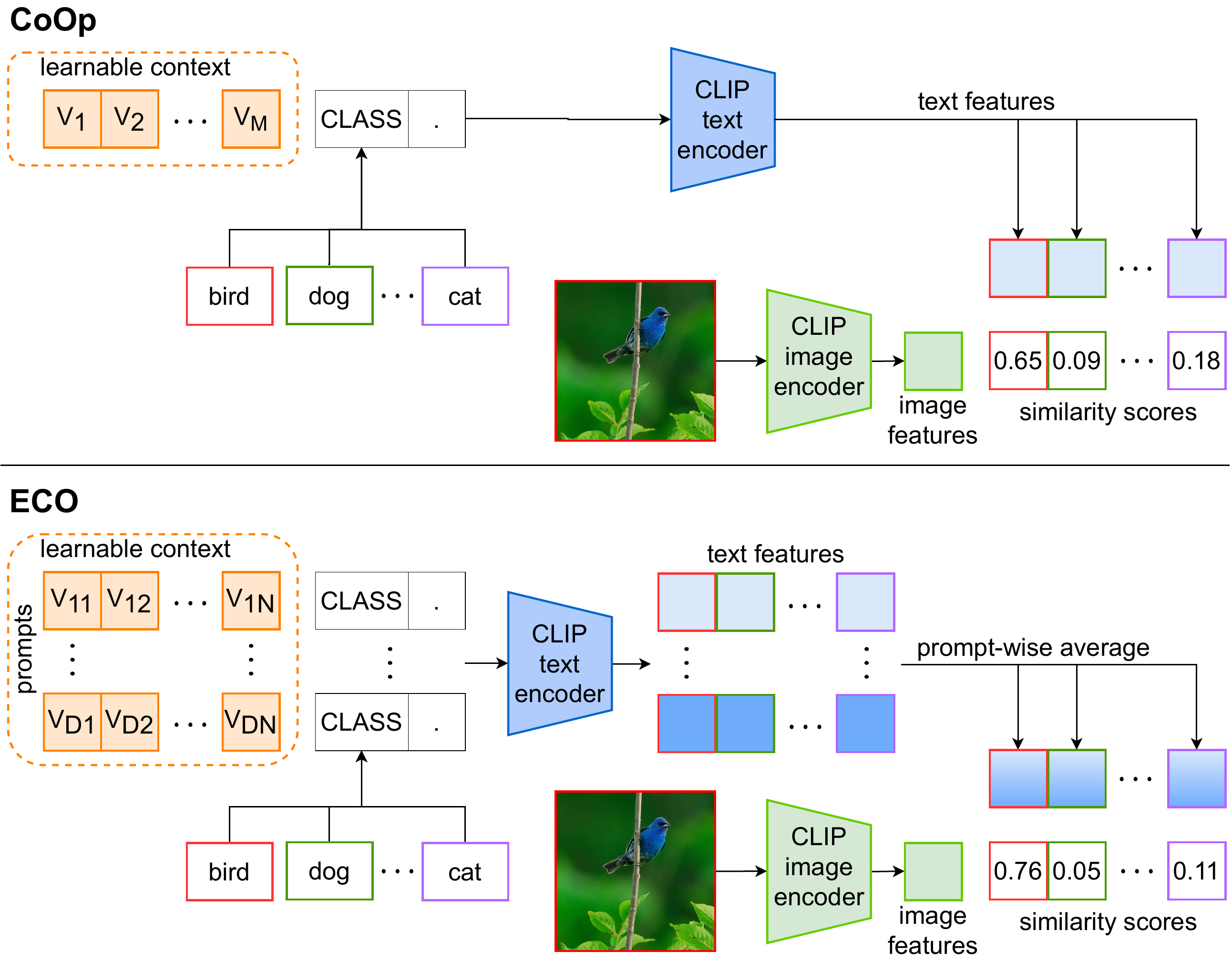}
    \caption{Overview of our approach. While CoOp uses a single prompt with $M$ context tokens, \method trains $D$ prompts with $N$ context tokens each, such that $M\!=\!D*N$. Given the same number of trainable parameters, ensembling multiple prompts with a reduced number of context tokens performs better than using a single prompt with a larger number of context tokens.}
    \label{fig:teaser}
\end{figure}

However, the textual input -- referred to as \textit{prompt} -- greatly influences the performance in downstream tasks. For example, \cite{zhou2022learning} reports a $5\%$ increase in accuracy by adding an ``\textit{a}" before the class token in the prompt ``\textit{a photo of [CLASS]}" for few-shot classification with the Caltech101 \cite{fei2004learning} dataset. Given the significant difference in performance caused by slight changes in wording, crafting prompts by hand to find the best-performing one is a non-trivial task. Therefore, \textit{prompt ensembling} is often employed to improve the robustness and achieve better results \cite{radford2021learning}. Prompt ensembling consists of computing the textual features of several different prompts, such as ``\textit{a photo of a [CLASS]}", ``\textit{an illustration of a [CLASS]}" etc., and then using the average of them for the downstream task.

Recently, several works have proposed to employ prompt learning to substitute hand-crafted prompts with learned context word vectors. CoOp \cite{zhou2022learning} was the first work to propose to use prompt learning for vision-language models, improving over hand-crafted prompts. CoCoOp \cite{zhou2022conditional} trains a neural network to generate an input-conditional token for each image. MaPLe \cite{khattak2023maple} proposes to learn a multi-modal prompt instead of a textual-only one. However, all existing methods only learn a single prompt, thus not exploiting the potential of prompt ensembling.

For this reason, we present \method (Ensembling Context Optimization), a method for merging prompt learning and prompt ensembling. The main idea of our approach is conceptually quite straightforward: learning multiple prompts with a reduced number of context tokens instead of a single one with a larger number of context tokens, and then combining them with prompt ensembling. \Cref{fig:teaser} shows an overview of the proposed method and a comparison with CoOp \cite{zhou2022learning}. Note that \method is orthogonal to the prompt learning technique being used, as it focuses on how to take full advantage of the information of the learned prompts rather than how to obtain it. 
Despite its apparent simplicity, our approach performs significantly better than the competing methods on 11 testing datasets. Moreover, it proves to be a more data-efficient and effective few-shot learner, since the largest gains in performance are observed for as few as 1 and 2 shots. Finally, ECO does not add any computational overhead at inference time since the textual features used for the classification can be precomputed.

We summarize the contributions of this work as follows:
\begin{itemize}
    \item We propose \method, an approach for prompt learning that employs prompt ensembling to combine multiple prompts with reduced learned context tokens;
    \item \method can be combined with any prompt learning strategy, making it a simple and versatile tool for improving accuracy with no overhead at inference time;
    \item We obtain significant improvements over the competing methods on 11 testing datasets, showing the effectiveness of our method.
\end{itemize}

% \section{Prompt Learning}
% \begin{itemize}
%     \item This section could serve as a sort of brief related work that also helps the reader to remember what Prompt Learning actually is
%     \item Also explain how the baseline model works
% \end{itemize}

\section{Method}
\subsection{Preliminaries}
%\paragraph{CLIP} 
The vision-language model CLIP~\cite{radford2021learning} is designed to align visual and textual data within a common embedding space. It consists of two encoders: a visual encoder denoted as $f_{\theta}$ and a text encoder represented as $g_{\phi}$. These encoders extract feature representations $f_{\theta}(I) \in \mathbb{R}^{d}$ and $g_{\phi}(E_w(Y)) \in \mathbb{R}^{d}$ from an input image $I$ and its corresponding text caption $Y$, respectively. Here, $d$ indicates the dimension of the CLIP embedding space, while $E_w$ represents the word-embedding layer, which maps each tokenized word in $Y$ to the token embedding space $\mathcal{W}$.
The primary objective of training the CLIP model is to ensure a high similarity between the feature representations of corresponding images and text, i.e. $f_{\theta}(I) \approx g_{\phi}(E_w(Y)))$.

In the zero-shot classification setup using CLIP, we start with an image $I$ and a set of text prompts $\{Y_i\}_{i=1}^K$, where $K$ represents the number of classes. Each text prompt $Y_i$ is of the form ``\textit{a photo of a [$\text{CLASS}_i$]}", with $\text{CLASS}_i$ denoting a specific class name, such as ``\textit{bird}", ``\textit{dog}", ``\textit{cat}", \etc. We then extract feature representations from the image and the text prompts using the CLIP encoders. The image feature representation is denoted as $\psi_I = f_{\theta}(I)$, while the text feature representation for each prompt is represented as $\psi_T^i = g_{\phi}(E_w(Y_i))$.
Finally, we can compute the prediction probability for each class as follows:

\begin{equation} \label{eq:prob}
p(y=i|I) = \frac{\exp( \cos(\psi_T^i, \psi_I)/\tau) }{\sum_{j=1}^K \exp( \cos(\psi_T^j, \psi_I)/\tau)},
\end{equation}

Here, $\tau$ is a temperature parameter that is learned during the training of the CLIP model, and $\cos(\cdot, \cdot)$ represents the cosine similarity between the image and text features.

\subsection{\method}
Our approach, named \method, aims to enhance the adaptability of frozen pre-trained CLIP models to downstream tasks by overcoming the inefficiency of hand-crafted prompts. Previous methods, such as CoOp~\cite{zhou2022learning}, CoCoOp~\cite{zhou2022conditional}, and MaPLe~\cite{khattak2023maple}, learn a single set of context tokens. On the contrary, drawing inspiration from prompt ensembling techniques that have proven to boost performance over using a single prompt~\cite{radford2021learning}, we learn multiple sets of context tokens. In other words, while standard prompt learning techniques learn only a single prompt, we learn multiple prompts that we combine together to improve performance.
%Note that our ensembling approach also significantly differs from~\cite{radford2021learning}, that uses handcrafted prompts.

We denote the multiple sets of context tokens (\ie the learnable prompts) as $\{v_{i1},\ldots, v_{iN}\}_{i=1}^D$, where each context vector $v_{ij}$ belongs to the CLIP token embedding space $\mathcal{W}$. Here, $N$ represents the number of context tokens per prompt, while $D$ is the total number of prompts.
% In our approach, we learn multiple sets of continuous context tokens denoted as $\{v_{i1},\ldots, v_{iN}\}_{i=1}^D$, where each context vector $v_{ij}$ belongs to the CLIP token embedding space $W$. Here, $N$ represents the number of context vectors per prompt, and $D$ is the total number of prompts. 
For the k-th class of a dataset, the inputs to the text encoder are defined as $\{v_{i1},\ldots, v_{iN}, c_k\}_{i=1}^D$, where $c_k = E_w([CLASS_k])$.
Similarly to CoOp, we share the same set of context vectors among all classes. We then extract the textual features using the textual encoder, averaging across prompts $\psi_T^k = \frac{1}{D}\sum_{i=1}^D{g_{\phi}(\{v_{i1},\ldots, v_{iN}, c_k\})}$.  Consequently, we can compute the probability $p(y=k| I)$ using \cref{eq:prob}.

The key innovation lies in our use of multiple prompts. We learn distinct sets of context vectors from data instead of relying on hand-crafted prompts like "\textit{a photo of a [CLASS]}". Intuitively, each prompt contributes to a diverse feature extraction process, and we effectively blend the prompt-specific features by performing an element-wise average. This prompt-wise average conceptually emulates prompt ensembling, known to enhance CLIP's zero-shot classification performance~\cite{radford2021learning}. However, unlike standard prompt ensembling with hand-created prompts, our method learns context vectors directly from the data.
To summarize, \method seamlessly combines the concepts of prompt learning and prompt ensembling, a novel combination not previously explored in vision-language tasks.

During training, we employ cross-entropy as the loss function, allowing the gradients to flow through the text encoder to update the weights of the context vectors. Importantly, the CLIP base model remains frozen throughout the entire training process.
To ensure a fair comparison with CoOp, we keep the number of trainable parameters constant. If CoOp uses $M$ context vectors, we set $N$ and $D$ such that $M\!=\!N * D$. Note that our method coincides with CoOP when $D\!=\!1$ and $N\!=\!M$.
Although in our experiments we extend the CoOp method, what we propose is a general framework that can be extended to all prompt learning techniques that learn a single set of context tokens.
In addition, ECO does not add any computational overhead at inference time. Despite learning multiple contexts, after training these are fixed and their encodings are averaged into a single latent vector $\psi_T^k$. Since $\psi_T^k$ does not depend on the input, it can be stored and used as a single prompt, requiring no additional computation compared to non-ensembling models like CoOp.

% Although in our experiments we extend the CoOp method, what we propose is a general framework that can be extended to all prompt learning techniques that learn a single set of context tokens.

\section{Experimental Results}

\begin{figure*}[t]
    \centering
    \begin{subfigure}[b]{0.33\linewidth}
        \centering
        \includegraphics[width=\linewidth]{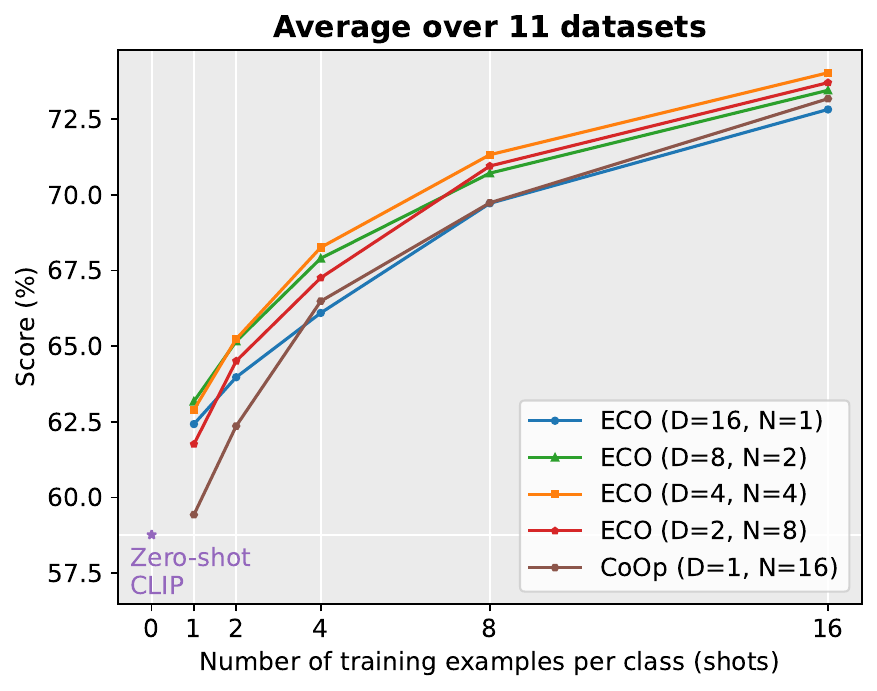}
    \end{subfigure}
    \hfill
    \begin{subfigure}[b]{0.33\linewidth}
        \centering
        \includegraphics[width=\linewidth]{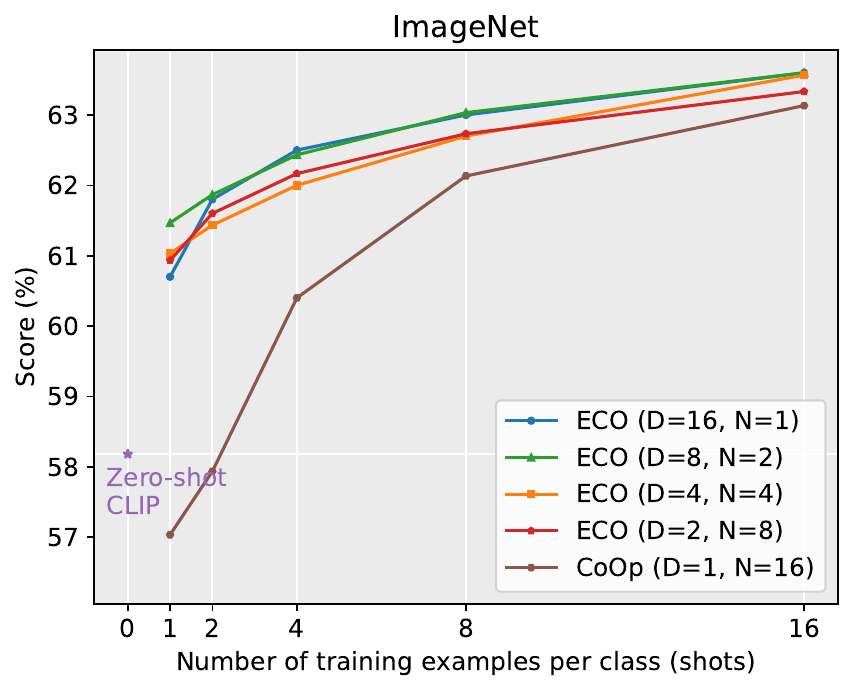}
    \end{subfigure}
    \hfill
    \begin{subfigure}[b]{0.33\linewidth}
        \centering
        \includegraphics[width=\linewidth]{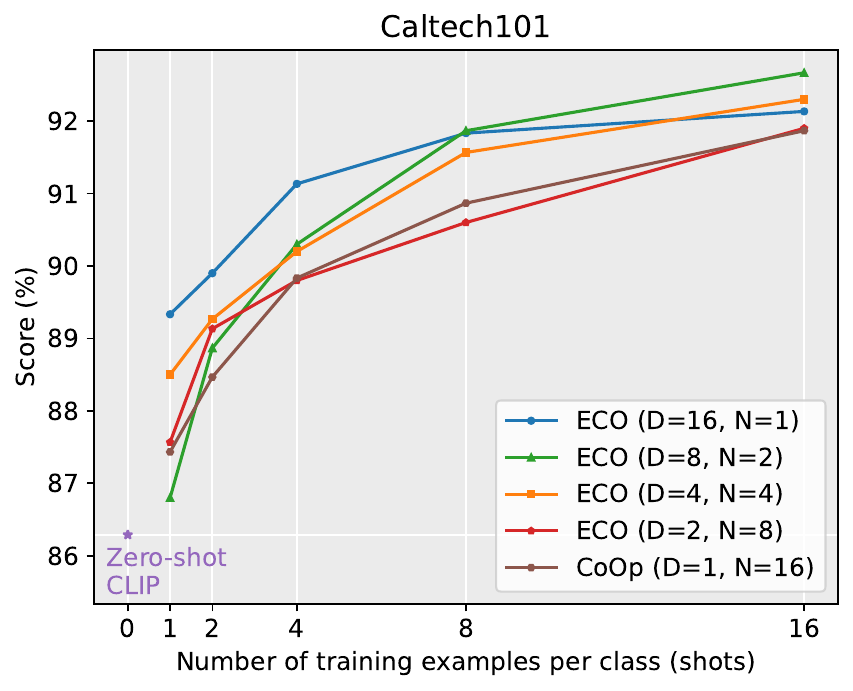}
    \end{subfigure}
    % \vspace{-1ex}
    
    \begin{subfigure}[b]{0.33\linewidth}
        \centering
        \includegraphics[width=\linewidth]{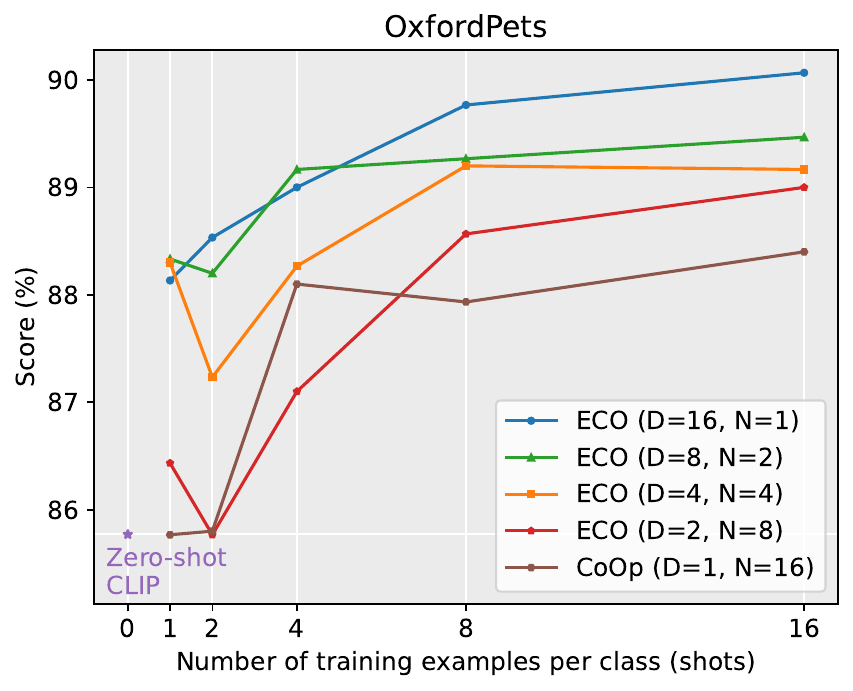}
    \end{subfigure}
    \hfill
    \begin{subfigure}[b]{0.33\linewidth}
        \centering
        \includegraphics[width=\linewidth]{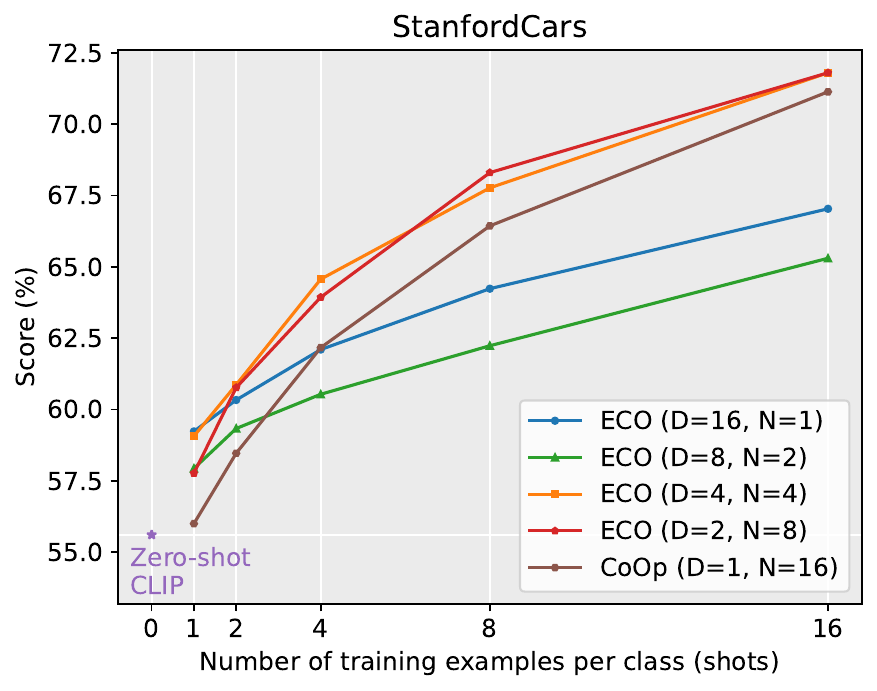}
    \end{subfigure}
    \hfill
    \begin{subfigure}[b]{0.33\linewidth}
        \centering
        \includegraphics[width=\linewidth]{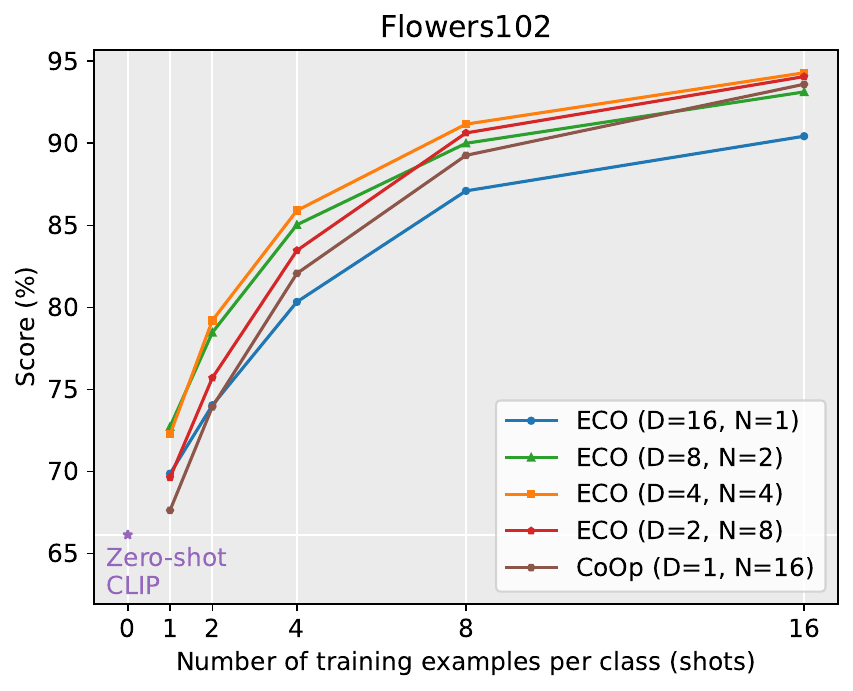}
    \end{subfigure}
    % \vspace{-1ex}
    
    \begin{subfigure}[b]{0.33\linewidth}
        \centering
        \includegraphics[width=\linewidth]{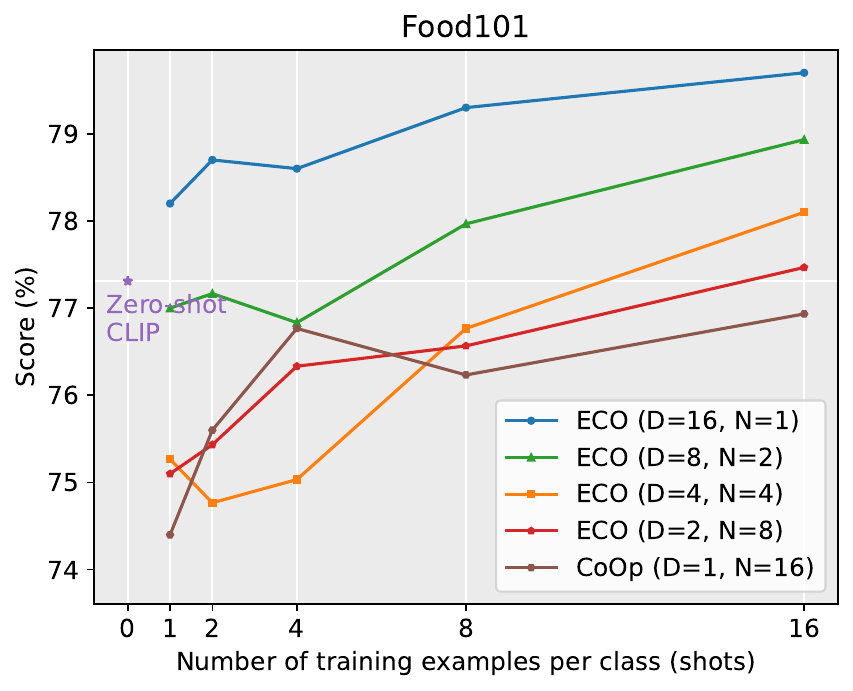}
    \end{subfigure}
    \hfill
    \begin{subfigure}[b]{0.33\linewidth}
        \centering
        \includegraphics[width=\linewidth]{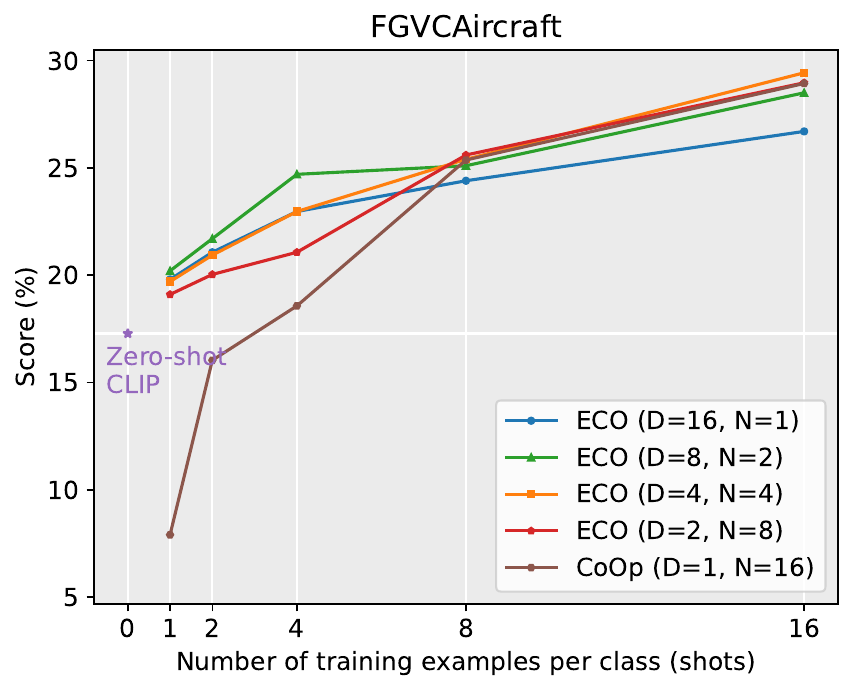}
    \end{subfigure}
    \hfill
    \begin{subfigure}[b]{0.33\linewidth}
        \centering
        \includegraphics[width=\linewidth]{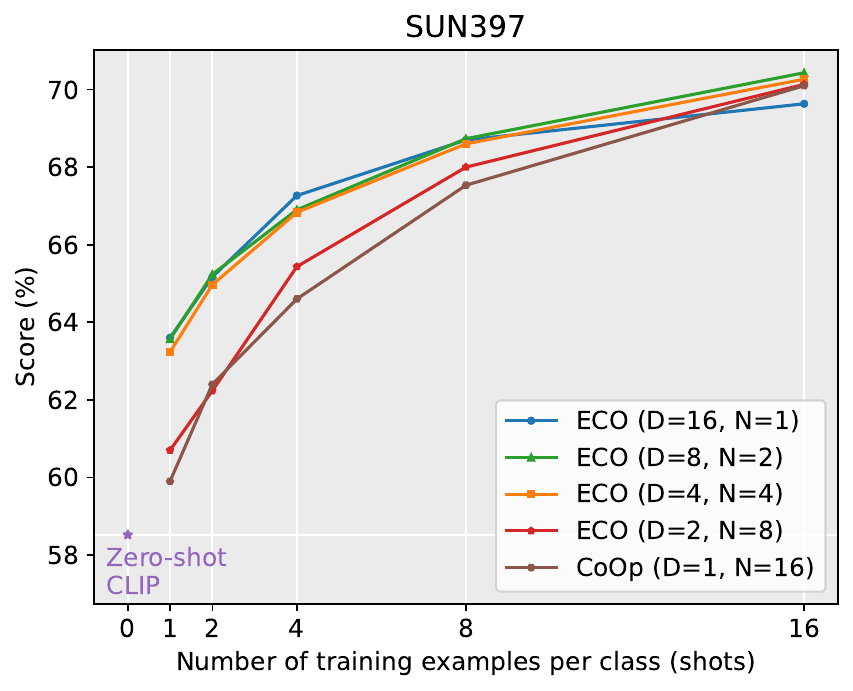}
    \end{subfigure}
    % \vspace{-1ex}
    
    \begin{subfigure}[b]{0.33\linewidth}
        \centering
        \includegraphics[width=\linewidth]{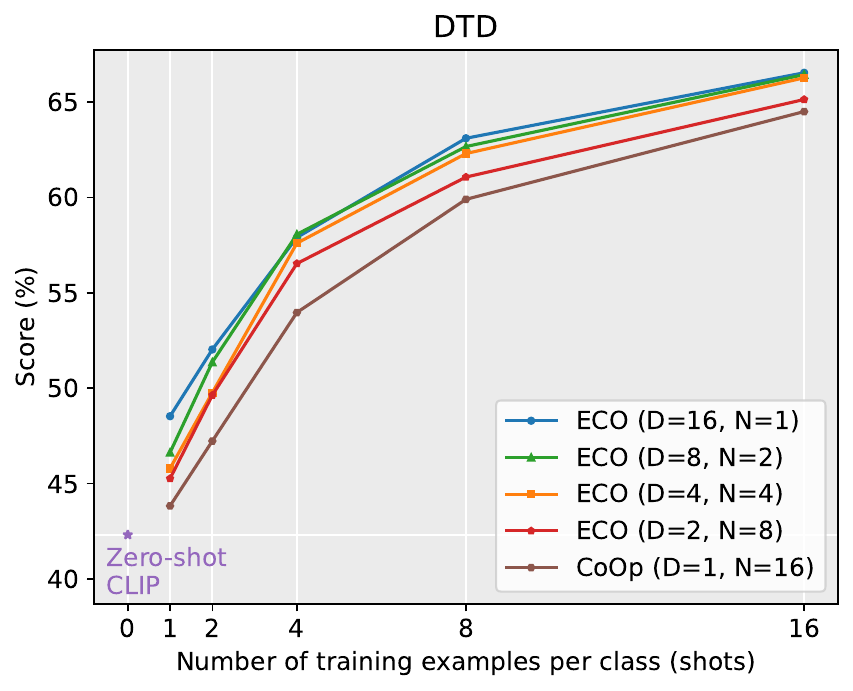}
    \end{subfigure}
    \hfill
    \begin{subfigure}[b]{0.33\linewidth}
        \centering
        \includegraphics[width=\linewidth]{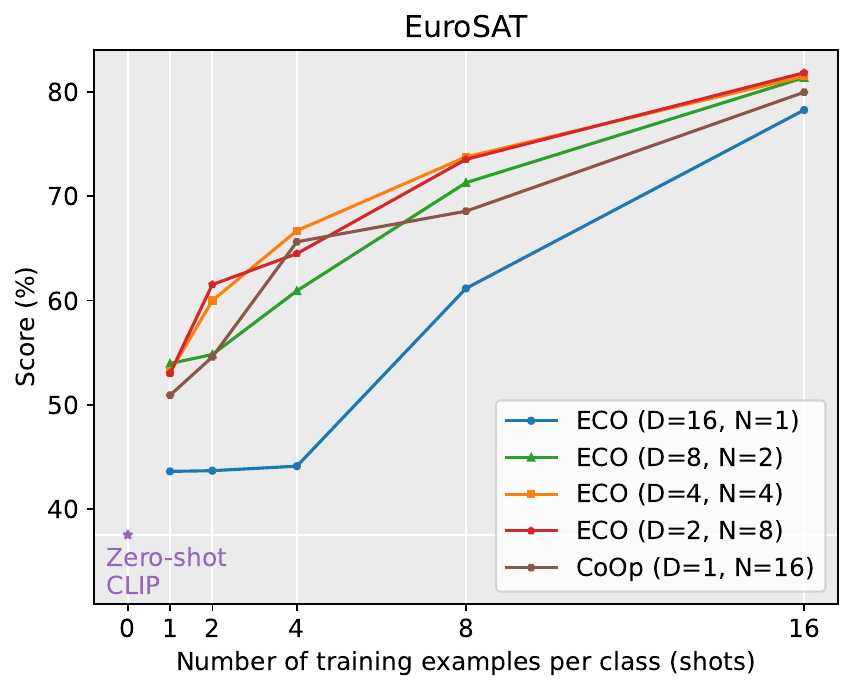}
    \end{subfigure}
    \hfill
    \begin{subfigure}[b]{0.33\linewidth}
        \centering
        \includegraphics[width=\linewidth]{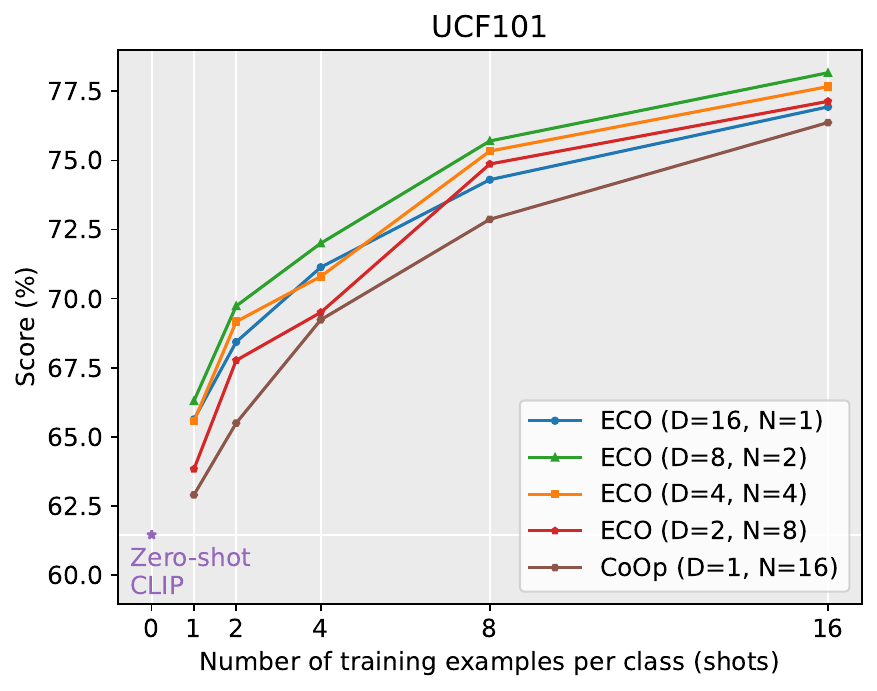}
    \end{subfigure}
    \caption{Quantitative results on the 11 test datasets varying the number of shots, prompts D and context tokens N for each of them. Note that CoOp~\cite{zhou2022learning} coincides with \method when $D\!=\!1$ and $N\!=\!16$.}
    \label{fig:quantitative_results}
\end{figure*}

\begin{table}[t]
  \centering
  \large
  \resizebox{\linewidth}{!}{ %< auto-adjusts font size to fill line
  \begin{tabular}{lccccc} 
  \toprule
  \multicolumn{1}{c}{} & \multicolumn{5}{c}{Shots} \\
  \cmidrule(lr){2-6}
  \multicolumn{1}{l}{Method} & $1$ & $2$ & $4$ & $8$ & $16$\\ 
  \midrule
  Zero-Shot CLIP$^{\ddagger}$~\cite{radford2021learning} & 58.77 & 58.77 & 58.77 & 58.77 & 58.77 \\ 
  Linear Probe CLIP~\cite{radford2021learning} & 36.67 & 47.61 & 57.19 & 64.98 & 71.10 \\
  CoOp~\cite{zhou2022learning} & 59.59 & 62.32 & 66.77 & 69.89 & 73.42 \\ \midrule[.02em]
  \method ($D\!=\!16$, $N\!=\!1$) & 62.42 & 63.97 & 66.10 & 69.72 & 72.82 \\ 
  \method ($D\!=\!8$, $N\!=\!2$) & \textbf{63.18} & 65.16 & 67.90 & 70.72 & 73.45 \\
  \method ($D\!=\!2$, $N\!=\!8$) & 61.76 & 64.51 & 67.26 & 70.95 & 73.71 \\ \midrule[.02em]
  CoOp$^{\dagger}$ ($D\!=\!1$, $N\!=\!16$)  & 59.43 & 62.36 & 66.49 & 69.74 & 73.18 \\ %\midrule[.02em]
  \rowcolor{tabhighlight}
  \method ($D\!=\!4$, $N\!=\!4$) & 62.90 & \textbf{65.24} & \textbf{68.26} & \textbf{71.33} & \textbf{74.03} \\
  & \textcolor{MidnightBlue}{{+3.47}} & \textcolor{MidnightBlue}{{+2.88}} & \textcolor{MidnightBlue}{{+1.77}} & \textcolor{MidnightBlue}{{+1.59}} & \textcolor{MidnightBlue}{{+0.85}} \\
  \bottomrule
  \end{tabular}}
  \caption{Detailed comparison of the results on the average of the 11 datasets. Best scores are highlighted in bold. $^{\ddagger}$ uses always zero shots. $^{\dagger}$ indicates results obtained with our implementation. Note that CoOp$^{\dagger}$ coincides with \method ($D\!=\!1$, $N\!=\!16$). Absolute gains over CoOp$^{\dagger}$ \cite{zhou2022learning} are indicated in \textcolor{MidnightBlue}{{blue}}.}
  \label{tab:quantitative_results}
\end{table}

Since \method does not depend on a specific prompt learning technique, we choose to compare our approach to the most basic one, \ie CoOp \cite{zhou2022learning}. In future work, we will extend the proposed method to other prompt learning works, such as CoCoOp \cite{zhou2022conditional} and MaPLe \cite{khattak2023maple}.

\subsection{Evaluation Protocol}
We follow the few-shot evaluation protocol of \cite{radford2021learning, zhou2022learning}, using 1, 2, 4, 8, and 16 shots for training and evaluating the performance of each model in the full test sets. We report the average results over three seeds.

Similarly to \cite{zhou2022learning}, we evaluate our approach on 11 image classification datasets: ImageNet \cite{deng2009imagenet}, Caltech101 \cite{fei2004learning}, OxfordPets \cite{parkhi2012cats}, StanfordCars \cite{krause20133d}, Flowers102 \cite{nilsback2008automated}, Food101 \cite{bossard2014food}, FGVCAircraft \cite{maji2013fine}, SUN397 \cite{xiao2010sun}, DTD \cite{cimpoi2014describing}, EuroSAT \cite{helber2019eurosat} and UCF101 \cite{soomro2012ucf101}.

We consider the version of CoOp with the class token positioned at the end, ResNet 50 \cite{he2016deep} as the backbone and with the number of context tokens $M\!=\!16$. We inherit the training details from \cite{zhou2022learning}. For a fair comparison, in the experiments, we vary the number of prompts $D$ and the number of context tokens $N$ for each of them so that the number of trainable parameters stays the same (\ie $M\!=\!D * N$). Note that, for $D\!=\!1$ and $N\!=\!16$, \method coincides with CoOp.

\subsection{Quantitative Results}
\Cref{fig:quantitative_results} shows the results of \method for all the testing datasets. For completeness, we also report the performance of zero-shot CLIP, which is based on hand-crafted prompts. \method obtains significant improvements over the baselines on all the datasets. The version of \method with $N\!=\!4$ and $D\!=\!4$ achieves the best performance on average and proves to be the best tradeoff between the number of prompts and context tokens. 
% Moreover, the results show that \method is a better few-shot learner and is more data-efficient than CoOp \cite{zhou2022learning} since the largest gains in performance are obtained for as few as 1 and 2 shots, \eg on ImageNet and FGVCAircraft.
In addition, \method is less sensitive to noisy labels than CoOp \cite{zhou2022learning}, as it achieves better performance than zero-shot CLIP also on the Food101 dataset, which is known to have noisy annotations~\cite{zhou2022learning}. 

In \Cref{tab:quantitative_results} we provide a comparison between the different versions of \method and the baselines by reporting the average accuracy on the 11 benchmarks, varying the number of shots. For a fair comparison, for CoOp we report the results we obtained with the version of our model with $D\!=\!1$ and $N\!=\!16$, since they coincide. We denote this in  \Cref{tab:quantitative_results} as CoOp$^{\dagger}$. However, we observed a difference of only $0.32\%$ on average with the values of the original paper \cite{zhou2022learning}, which can be attributed to different seeds and hardware. For completeness, we also provide the results of the linear probe model of CLIP, which is considered a strong few-shot learning baseline \cite{tian2020rethinking}. Our approach consistently outperforms all the competing methods. In particular, we observe absolute improvements up to 3.47 over CoOp. Moreover, the results show that \method is a better few-shot learner and is more data-efficient than CoOp since the largest gains in performance are obtained for as few as 1 and 2 shots.

Overall, the experimental results demonstrate that, when utilizing an equivalent number of trainable parameters, employing an ensemble of multiple prompts with a reduced number of context tokens performs better than using a single prompt with a larger number of context tokens.

% In \Cref{tab:quantitative_results} we provide a comparison between the best-performing version of \method and CoOp by reporting the average accuracy on the 11 benchmarks, varying the number of shots. 
% For a fair comparison, for CoOp we report the results we obtained with the version of our model with $D\!=\!1$ and $N\!=\!16$, since they coincide. We denote this in  \Cref{tab:quantitative_results} as CoOp$^{\dagger}$. However, we observed a difference of only $0.32\%$ on average with the values of the original paper \cite{zhou2022learning}, which can be attributed to different seeds and hardware.
% For completeness, we also provide the results of the linear probe model of CLIP, which is considered a strong few-shot learning baseline \cite{tian2020rethinking}. Our approach consistently outperforms CoOp, obtaining absolute improvements up to 3.47 for the single shot configuration.

% We also report in \Cref{tab:quantitative_results} the average results for different combinations of $D$ and $N$, as in \Cref{fig:quantitative_results}.
% Overall, the experimental results demonstrate that, when utilizing an equivalent number of trainable parameters, employing an ensemble of multiple prompts with a reduced number of context tokens performs better than using a single prompt with a larger number of context tokens. 

\section{Conclusion}
In this paper, we have proposed a novel prompt learning strategy that consists in optimizing an ensemble of multiple contexts. Although simple, the method is effective, yielding consistent improvements over 11 different benchmarks, and versatile, being it applicable on top of potentially any existing prompt learning technique with no additional overhead at inference time.
Interestingly, we found that balancing context length and number of prompts is beneficial for effectively exploit CLIP for few-shot image classification. This is particularly true for a reduced number of shots, such as 1 or 2, for which we report the bigger gains.

{\small
\bibliographystyle{ieee_fullname}
\bibliography{egbib}
}

\end{document}